\documentclass[conference]{IEEEtran}
\usepackage{times}

\usepackage{bm}
\newcommand\blfootnote[1]{%
  \begingroup
  \renewcommand\thefootnote{}\footnote{#1}%
  \addtocounter{footnote}{-1}%
  \endgroup
}

\usepackage{graphicx}                       % Necessary to include graphics
\usepackage{graphics}                       % To include pdf, bitmapped graphics files
\usepackage{epsfig}                         % To include eps files
\usepackage[tight,footnotesize]{subfigure}  % Create subfigures, ie 1A, 1B
\graphicspath{./pics/RAL_fig_for_revise}
\graphicspath{./pics/RAL_new_fig}
\usepackage{amssymb,amsmath}%\usepackage{amslatex}%
\usepackage{mdwmath}
\usepackage{commath}   % Used for \norm \abs
\usepackage{eqparbox}
\usepackage{mathtools}
\usepackage[utf8]{inputenc} % Used for theorem/corollary/lemma
\usepackage[english]{babel}
\usepackage{amsmath,amsfonts,amssymb}

\newcommand{\beq}{\begin{equation}}
\newcommand{\eeq}{\end{equation}}
\newcommand{\bear}{\begin{eqnarray}}
\newcommand{\bears}{\begin{eqnarray*}}
\newcommand{\eear}{\end{eqnarray}}
\newcommand{\eears}{\end{eqnarray*}}
\newcommand{\bdm}{\begin{displaymath}}
\newcommand{\edm}{\end{displaymath}}
\newcommand{\lba}{\left[\begin{array}}
\newcommand{\ear}{\end{array}\right]}

\usepackage{stfloats}                       % Create Navigation Titles in PDF
\usepackage{url} % bibtex
\usepackage{hyperref}
\usepackage{cite}                           % Allows to establish ranges of papers for extended citations.
\usepackage[T1]{fontenc} % Underscores (to better visualize them using the appropriate font)
\usepackage{tabularx}
\usepackage{multirow}
\usepackage[table]{xcolor}
\usepackage{longtable}
\usepackage{booktabs} 			% professional-quality tables and \midrule
\usepackage{xcolor,colortbl}    % Color columns, use with: \usepackage{array}

\usepackage{nicematrix}
\usepackage{multicol}
\usepackage{graphicx}
\usepackage{placeins}
\usepackage{multirow}

\begin{document}

%----------------------------------------------------------------------------------

\title{Probabilistic Discriminative Models Address\\ the Tactile Perceptual Aliasing Problem}

%----------------------------------------------------------------------------------\dagger This is to avoid two footnotes numbered by 1

\author{
\authorblockN{John Lloyd, Yijiong Lin and Nathan F. Lepora\\ 
Department  of  Engineering  Mathematics  and  Bristol  Robotics  Laboratory,  University  of  Bristol,  U.K.\\
Email: \{jl15313, yijiong.lin, n.lepora\}@bristol.ac.uk}
}
% \authorblockN{Author Names Omitted for Anonymous Review \vspace{1cm}}
%}

\maketitle

%\IEEEpeerreviewmaketitle

\begin{abstract}
In this paper, our aim is to highlight {\em Tactile Perceptual Aliasing} as a problem when using deep neural networks and other discriminative models. Perceptual aliasing will arise wherever a physical variable extracted  from  tactile data is subject to ambiguity between stimuli that are physically distinct. Here we address this problem using a probabilistic discriminative model implemented as a 5-component mixture density network comprised of a deep neural network that predicts the parameters of a Gaussian mixture model. We show that discriminative regression models such as deep neural networks and Gaussian process regression perform poorly on aliased data, only making accurate predictions  when the sources of aliasing are removed. In contrast, the mixture density network identifies aliased data with improved prediction accuracy. The uncertain predictions of the model form patterns that are consistent with the various sources of perceptual ambiguity. In our view, perceptual aliasing will become an unavoidable issue for robot touch as the field progresses to training robots that act in uncertain and unstructured environments, such as with deep reinforcement learning.\blfootnote{This research was supported in part by a Leverhulme Trust Research Leadership award (grant number RL-2016-39 to NL).}
\end{abstract}
%----------------------------------------------------------------------------------
\section{INTRODUCTION}\label{sec:Intro}
%----------------------------------------------------------------------------------

For half a century, tactile sensing has held promise to bring human-like dexterity to robot hands and manipulators. However, there has been some degree of over-optimism about how quickly this change will take place. For example, a seminal review of Automated Tactile Sensing from 1982 concluded~\cite{harmon_automated_1982}: "Intelligent, dextrous, competent, manipulating robots will surely make their appearance and pay their way in society within the coming decade." There are various reasons why this scenario still seems out of reach nearly four decades later~\cite{billard_trends_2019,yuan_touch_2020}; however, the deep learning revolution~\cite{lecun_deep_2015} that has swept through computer vision~\cite{krizhevsky_imagenet_2017} and is now sweeping through robotics~\cite{sunderhauf_limits_2018} gives a renewed feeling of optimism that intelligent, dexterous robots will materialise this decade. That said, tactile sensing differs fundamentally from vision~\cite{hayward_is_2011}. While there may be some cross-over of progress between the two modalities, tactile sensing has its own unique challenges.

\begin{figure}[t!]
  \centering
    \includegraphics[width=\linewidth]{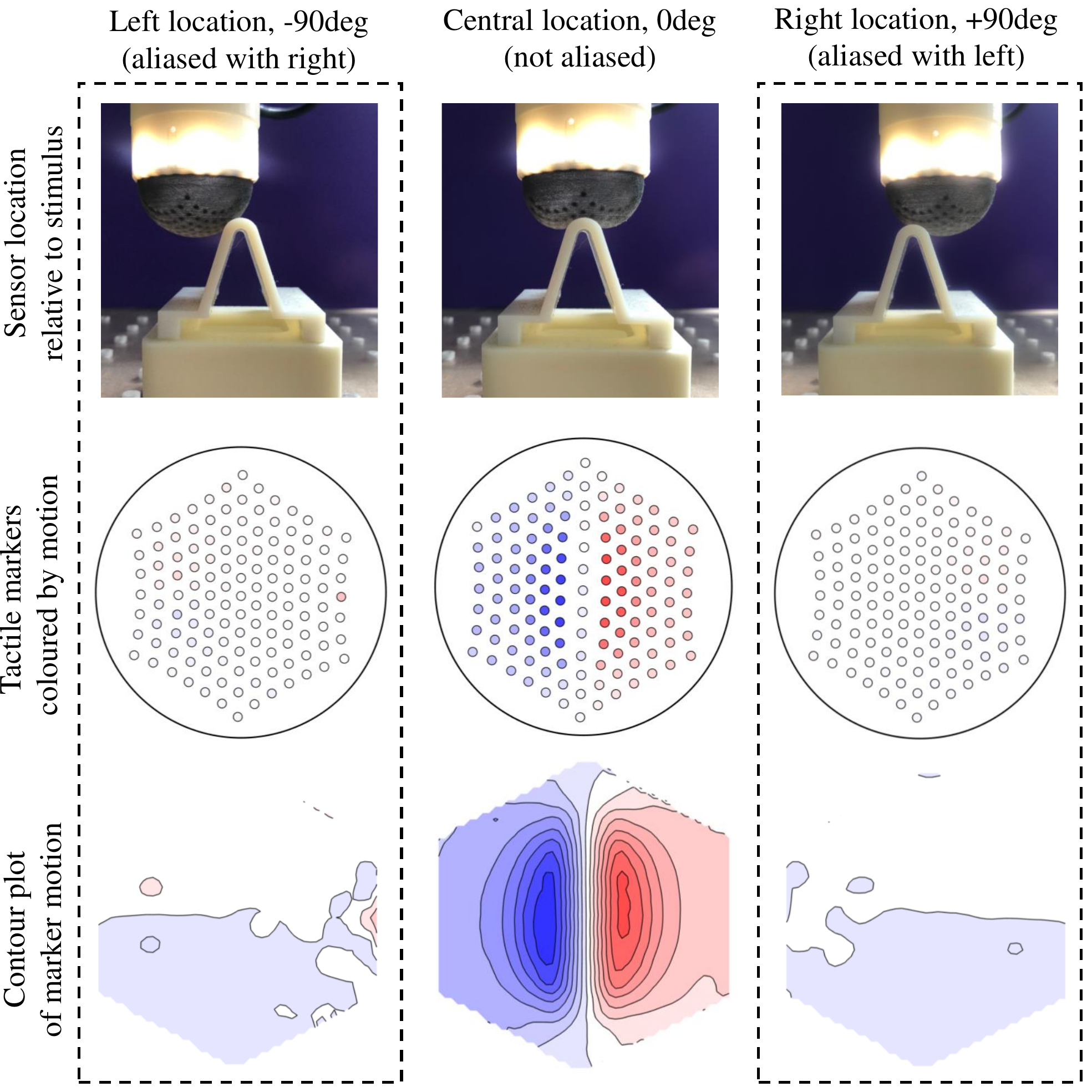}
    \caption{Tactile perceptual aliasing. The tactile sensor contacts a ridged stimulus from 3 locations (the sensor is rotated by -90\,deg, 0\,deg and +90\,deg relative to the central location in moving between the three contacts). Markers are coloured by $x$-displacement for visualisation. The peripheral contacts (left, right) produce a similar sensor output even though their locations are very different, and they are both distinct from the central contact. Note that even without rotating the sensor, discriminative models can find it difficult to distinguish the weak left and right contacts from each other. }
  \label{fig:perception_task}
  \vspace{-1em}
\end{figure}

One such challenge is the {\em Tactile Perceptual Aliasing Problem} (Fig.~\ref{fig:perception_task}). Aliasing is a well-known issue in vision when there is ambiguity in perceiving the stimulus~\cite{williams_aliasing_1985}, and this is known to cause difficulties for AI algorithms such as reinforcement learning~\cite{whitehead1991learning,chrisman_reinforcement_1992}. To the best of our knowledge, perceptual aliasing has received no attention in the artificial tactile sensing literature. The closest body of knowledge is on the subject of tactile illusions~\cite{de_vignemont_bodily_2005,hayward_brief_2008}, where it was observed~\cite{hayward_tactile_2015}: "One source of tactile illusion is clearly derived from contact mechanics... extracting the attributes of a touched object from partial knowledge of one's own tissue deformation, is a noisy and ambiguous process... and is at the root of all [tactile illusions] described thus far."

In this paper, our aim is to highlight {\em Tactile Perceptual Aliasing} as a problem when using deep neural networks and other discriminative models such as Gaussian process regression. The underlying cause of this problem is that tactile sensory data can be limited in its ability to disambiguate stimuli that may be physically distinct (see Fig.~\ref{fig:perception_task}). Camera-based optical tactile sensors such as the one used in this study, offer higher-resolution tactile images than conventional taxel arrays~\cite{shimonomura_tactile_2019}, but as we will see they are still subject to significant perceptual aliasing. In our view, aliasing is a generic problem for artificial tactile sensing that will become more prevalent as applications move from idealised research studies to practical robotics. Perceptual aliasing will arise whenever a physical variable extracted from a tactile image is subject to ambiguity; for example, as a contact gets lighter, the tactile imprint becomes less defined and therefore more ambiguous to localize on the sensor surface.

In future, we expect that multiple ways of dealing with tactile perceptual aliasing will be developed and used. One method that already appears to be widely used involves restricting the range over which the tactile data is collected so that ambiguous examples are excluded. Such data exclusion techniques are common in computer vision; for example, images were only selected for ImageNet if multiple human labellers agreed with each other~\cite{deng_imagenet_2009}. For tactile perception of physical quantities such as edge location and orientation, recent work with deep neural networks has used training ranges that implicitly have enough contact strength to be unambiguous ({\em e.g.} \cite[Table 1]{lepora_optimal_2020} excludes depths $<$1\,mm). Clearly, there are many issues with the data selection method: one is the need for human intervention to fine-tune the data collection; another is the difficulty in defining suitable ranges based on static conditions of soft, compliant sensors or  objects. 

Here we present an alternative way of addressing tactile perceptual aliasing using a probabilistic discriminative model that captures the distribution of the training data. The model produces both a prediction and an associated uncertainty based on the modelled distribution. It also explicitly models the distribution, and hence is more informative than methods that only produce single-point estimates of uncertainty.  Regions of ambiguity in the experimental parameter space indicate underlying sources of perceptual aliasing. The specific probabilistic discriminative model that we use here relies on a deep neural network to predict the parameters of a Gaussian mixture model, resulting in an architecture known as a Mixture Density Network (MDN)~\cite{bishop2006pattern}. 

The main contribution of this work is to highlight the problem of perceptual aliasing in tactile sensing, and to show how probabilistic discriminative models can address this issue by estimating prediction uncertainty and capturing the underlying probability distributions.  More specifically:\\
\noindent 1) We consider tactile perception of position, orientation and curvature under experimental conditions that introduce various sources of perceptual aliasing ({\em e.g.} light-contact data). Two data sets of 9,000 tapping contacts that span the range of tactile properties are used in this study. \\
\noindent 2) We apply two popular discriminitive regression methods to this data: Gaussian process regression and deep neural networks.  All model hyperparameters are tuned using Bayesian optimization to ensure a fair comparison. \\
\noindent 3) We also apply a probabilistic discriminative method to this data: a 5-component mixture density network. Once again, the hyperparameters are tuned using Bayesian optimization. \\
\noindent 4) We show that the discriminative regression models perform poorly on aliased data, only predicting accurately when the sources of aliasing are removed. In contrast, the MDN generates more accurate predictions and can identify aliased data.

%--------------------------------------------------------    

%----------------------------------------------------------------------------------
% \vfill
% \pagebreak

%----------------------------------------------------------------------------------
\section{BACKGROUND} \label{sec:background}
%----------------------------------------------------------------------------------
%----------------------------------------------------------------------------------
%\subsection{Generative vs discriminative models}\label{subsec:gen_vs_dis_model}
%----------------------------------------------------------------------------------
%Latent variables can also be incorporated if required. 

Models in tactile sensing are typically thought of in terms of the class of algorithm used or the type of information they produce; for example, a random forest classifier or support vector machine for classifying slip~\cite{veiga_stabilizing_2015,james2018slip} or a convolutional neural network for predicting shape-independent hardness~\cite{yuan_shape-independent_2017} or grasp success~\cite{calandra_feeling_2017}. However, from a statistical perspective, we believe there are more informative ways of viewing these models: whether they are generative or discriminative~\cite{ulusoy2005generative,jebara2012machine}, and the degree to which they explicitly or implicitly model the underlying probability distribution.%, and various flavours thereof.  

\textit{Generative models} capture the full joint probability density function (PDF) $p(\boldsymbol{x},\boldsymbol{y})$ over sensor output $\boldsymbol{x}$ and tactile stimulus $\boldsymbol{y}$. The distribution is usually represented as a product of the marginal and conditional PDFs $p(\boldsymbol{x})$ and $p(\boldsymbol{y}|\boldsymbol{x})$, or $p(\boldsymbol{y})$ and $p(\boldsymbol{x}|\boldsymbol{y})$. This type of model is referred to as “generative” because knowledge of the joint distribution is sufficient to generate samples from the model. Once the joint distribution is known, inference can be performed using probabilistic operations such as marginalization, conditioning or Bayes’ rule. However, this often requires large amounts of data, or strong assumptions such as the conditional independence assumption made in the naïve Bayes classifier. Examples of generative models used in tactile sensing have been based mainly on naive Bayes and related histogram likelihood models~\cite{jamali_majority_2011,liu_computationally_2012,lepora_active_2013,fishel_bayesian_2012}. 

\textit{Discriminative models}, on the other hand, only capture the conditional PDF or some underlying statistical properties of the distribution PDF~\cite{ulusoy2005generative}. This typically makes them more data-efficient, faster to train, and often more accurate, as long as it is possible to obtain enough labelled examples~\cite{jebara2012machine}.  Consequently, this type of approach has been used far more widely than generative approaches in tactile information processing. Examples include most of the classification and regression methods that are currently in use, {\em e.g.}~\cite{veiga_stabilizing_2015,james2018slip,yuan_shape-independent_2017,calandra_feeling_2017}, and many more covered in reviews such as~\cite{luo_robotic_2017}.

One can also distinguish between discriminative approaches that explicitly model the  conditional PDF ({probabilistic discriminative models}) and discriminative approaches that model useful properties of an implicit underlying PDF ({discriminative regression or classification models}). 

In this study, we show that for tactile perception there are benefits to be gained from explicitly modelling a more flexible form of conditional output distribution as a PDF and predicting information using that distribution. This contrasts with many other contemporary approaches that assume (often implicitly) a simplified output distribution such as a Gaussian.

Here we model conditional PDFs using mixture density networks. Other approaches for modelling conditional PDFs were also considered, including kernel-based models~\cite{sugiyama2010least}, conditional variational autoencoders (VAE)~\cite{sohn2015learning}, conditional GANs~\cite{mirza2014conditional}, normalizing flows~\cite{rezende2015variational} and neural autoregressive methods that factorize the output distribution~\cite{uria2016neural}.  However, like many other non-parametric approaches, kernel-based methods tend not to scale well to the large data sets that are now common in tactile sensing. Moreover, while other more complex `conditional-generative' methods are good at modelling the underlying data generation process, particularly for high-dimensional spaces, they tend to require computationally-expensive numerical integration or optimization to extract the associated PDF or related statistics.

%----------------------------------------------------------------------------------

%----------------------------------------------------------------------------------
\section{METHODS} \label{sec:method}
%----------------------------------------------------------------------------------

\subsection{Preliminary} \label{subsec:preliminary}

We model the output of the tactile sensor as a continuous random vector $\boldsymbol{x}=(x_1,x_2,...,x_M) $ that represents the output of individual tactile elements, or `taxels'. We also model a tactile stimulus as a continuous random vector $\boldsymbol{y}=(y_1,y_2,...,y_N) $ that represents attributes such as position, orientation and curvature. The probabilistic relationship between stimulus and sensor output is governed by a joint probability density function (PDF) $p(\boldsymbol{x},\boldsymbol{y})$.  Therefore, the inference process of tactile perception, where the applied stimulus that gave rise to an observed sensor output is inferred, is modelled by the conditional PDF $p(\boldsymbol{y}\mid \boldsymbol{x})$. Details of this formalism are given in the Appendix, with summaries of the model choice and their configurations presented below. 

\subsection{Model Configurations} \label{subsec:model_config}

In this study, we compare the performance of two popular discriminative regression models (Gaussian processes and neural networks) to a probabilistic discriminative model (mixture density networks).  These regression models were selected because of their widespread popularity and because they have some interesting contrasting properties.

\subsubsection{Gaussian Processes} \label{subsubsec:GP_config}
The kernel type and length scale were optimized in an outer loop using Bayesian optimization~\cite{snoek2012practical}; parameters including the signal and noise levels were optimized in an inner loop using the built-in quasi-Newton optimizer.  We tried several different types of kernel, including the exponential, Matern 3/2, Matern 5/2, rational quadratic and squared exponential kernels~\cite{10.5555/1162254}.  All other parameters and settings were left at their default values.
To cope with the relatively large datasets used in this study, we experimented with several different sparse approximation methods, including Subset of Data (SoD), Subset of Regressors (SoR) and Fully-Independent Conditional (FIC) methods~\cite{quinonero2005unifying}.  Of these, the SoR method consistently produced the best performance, and so we used it for all of our GP models.

\subsubsection{Deep Neural Networks} \label{subsubsec:NN_config}
The output layer is a dense (fully-connected) layer with linear activation function.  All hidden layers comprise four sub-layers: dense, followed by batch normalization, activation function, then dropout.  We experimented with several activation functions and obtained the best performance using Exponential Linear Units (ELUs)~\cite{clevert2015fast}.

To improve generalization, we optimized the dropout and L2 regularization coefficients, together with the number of hidden layers and hidden layer size, in an outer loop using Bayesian optimization~\cite{snoek2012practical}.  We found that dropout was most effective using a coefficient in the range 0 to 0.5.  We applied an L2-norm regularizer to the hidden- and output-layer weights, and to the gamma parameter of the batch normalization~\cite{ioffe2015batch} sub-layers.  This regularization was found to be most effective with a coefficient in the range 0.1 to 10.0.

All neural network models were trained using the Adam optimizer~\cite{kingma2014adam} for 2,000 training epochs with a batch size of 256.  For all models, we used a learning rate of 0.001 with an exponential decay rate of 0.5 over 10,000 decay steps.

\subsubsection{Mixture Density Networks} \label{subsubsec:MDN_config}
\begin{figure}[t]
  \centering
    \includegraphics[width=1\linewidth]{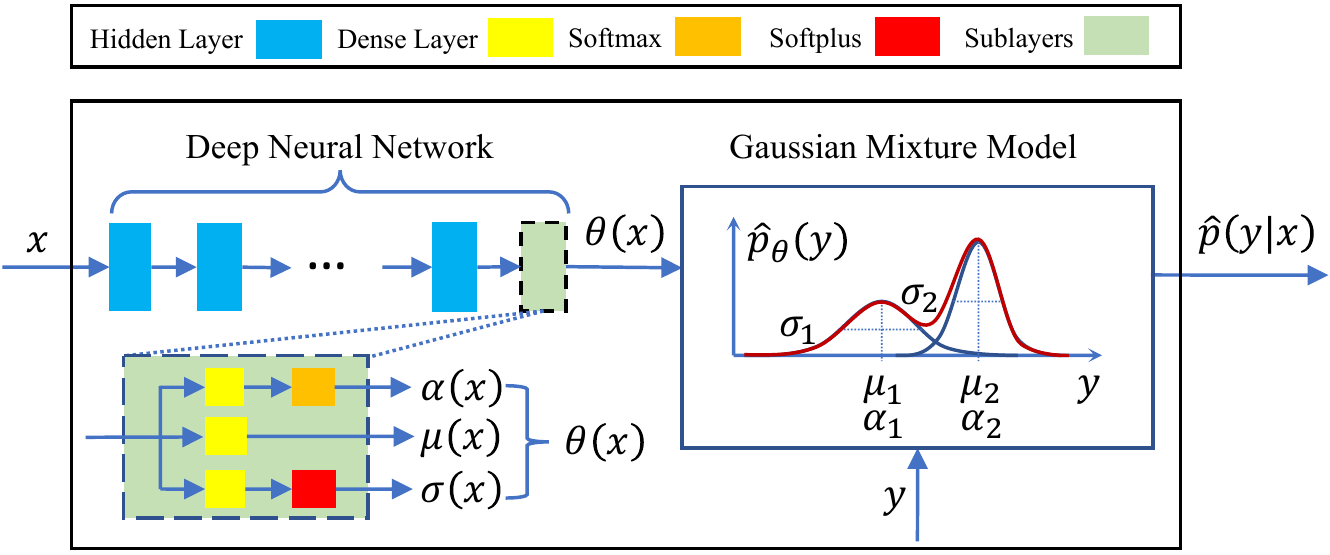}
    \caption{The mixture density network used here relies on a deep neural network to predict the parameters of a Gaussian mixure model. The neural network output layer is comprised of three parallel sub-layers (each having a different internal structure) for predicting the three different types of mixture model parameters (component weights, means and standard deviations).
    }
  \label{fig:MDN_structure}
\end{figure}

In this study, we focus on predicting isolated tactile properties so a univariate Gaussian mixture model is used to model the output distributions.  In this context, a deep neural network is used to predict the mixture weights and scalar-valued component means and standard deviations, based on the network inputs (Fig. \ref{fig:MDN_structure}).

The MDN output layer has three sub-layers, corresponding to the three different types of GMM parameter: a) the component means $\boldsymbol{\mu}_{i}$ are implemented using a dense layer, as for the neural network regression model; b) the mixture weights $\boldsymbol{\alpha}_{i}$ are implemented using a dense layer, followed by a \textit{softmax} activation function that ensures the corresponding outputs are positive and sum to one. The standard deviations $\boldsymbol{\sigma}_i$ are implemented by a dense layer, followed by the \textit{softplus} activation function, $f(x)=\log(1+\exp(x))$, which ensures that the corresponding outputs are positive.  We found that the \textit{softplus} activation function tends to produce more stable training than more nonlinear alternatives ({\em e.g.} the \textit{exp} function).

As for the neural network regression models, we applied batch normalization and dropout to all hidden layers.  We also applied an L2-norm regularizer to the hidden and output layer weights, and to the gamma parameter of the batch normalization sub-layers, using a scale parameter in the range 0.1 to 10.0. We found that higher levels of regularization were even more beneficial for MDN models (in terms of convergence speed and stability) than for neural network regression models, possibly because of the highly nonlinear way the predicted distribution parameters interact in the loss function.

During our initial experiments, we found no significant variation in predictive performance when using different numbers of mixture components, so long as there were enough components to represent salient features of the output distribution, such as multimodality. For the type of problems considered here, we found that a five-component MDN (which we denote as MDN5) provided an effective compromise between distribution fit, data requirements and training time. All MDNs were trained using the same type of optimizer and the same default training parameters that were used to train the neural network regression models.
 
%--------------------------------------------------------    
%-------------------------------------------------------- 

\subsection{Hyperparameter Optimization} \label{subsec:model_config}

The discriminative models used in this study have several hyperparameters that affect their prediction and generalization abilities. Typically, these hyperparameters are found using simple methods such as trial-and-error, grid search, random search, or more sophisticated search algorithms such as Bayesian optimization~\cite{snoek2012practical}.  In this study, we used Bayesian optimization to optimize the most important hyperparameters for the GP, NN and MDN models (Table \ref{tab:table_1}) and set the remaining ones to hand-optimized or default values.

The Bayesian optimizer was configured to use the `expected-improvement-plus' acquisition function, and run for 40 trial evaluations for each model.  For each Bayesian optimization trial, the models were trained on 80\% of the training data and evaluated on the remaining 20\%.  After the optimization process was complete, the optimized models were retrained using all of the training data.

We used the mean-squared error (MSE) between the predicted and target outputs of the validation data to evaluate the performance of the GPs and NNs, and the (conditional) negative log-likelihood of the validation data for the MDNs.  During our initial experiments, we found that certain hyperparameter combinations produced particularly large and noisy error values, which led to instability and inaccuracy in the optimization process, causing it to take longer or to fail catastrophically.  By way of mitigation, we found that applying the {\em logmod} function, $f(x)={\rm sign}(x)\cdot \log(1+|x|)$, to the error function helps compress extremely large and small values to reduce these problems. Since {\em logmod} is a monotonically-increasing function, the optima of the transformed loss lie at the same values as the original loss.

\begin{table} [b]
\centering 
    \caption{Model hyperparameters for optimization}
    \label{tab:table_1}
    \includegraphics[width=0.9\linewidth]{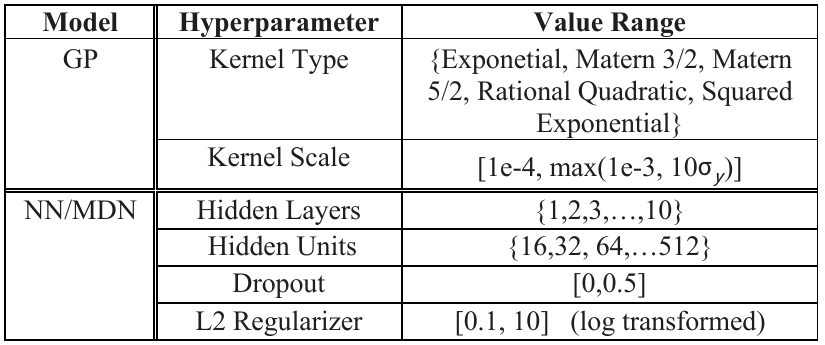}
\end{table}

\subsection{Robotic System and Software Environment} \label{subsec:hardware_software}

The robotic system used for this study consisted of an optical-based tactile fingertip sensor (the TacTip), mounted as an end-effector on an IRB 120 robot arm (ABB Robotics).

\subsubsection{Tactile Fingertip} \label{subsubsec:tactip}

We gathered tactile data using a TacTip optical tactile sensor developed at Bristol Robotics Laboratory that has since been distributed to other laboratories~\cite{ward2018tactip,lepora_soft_2021}. The 3D-printed sensor used in this paper has a hemispherical, gel-filled rubber tip with 127 internal marker pins, arranged in a hexagonal array with a pin spacing of approximately 3\,mm. The markers are illuminated by an LED ring situated around the circumference of the tip. When a mechanical stimulus is applied to the tip, the pins are displaced within the gel, and this movement is captured as a sequence of 680$\times$480-pixel images sampled at approximately 30\,fps.

%--------------------------------------------------------    

\begin{figure}[t]
  \centering
    \includegraphics[width=0.62\linewidth]{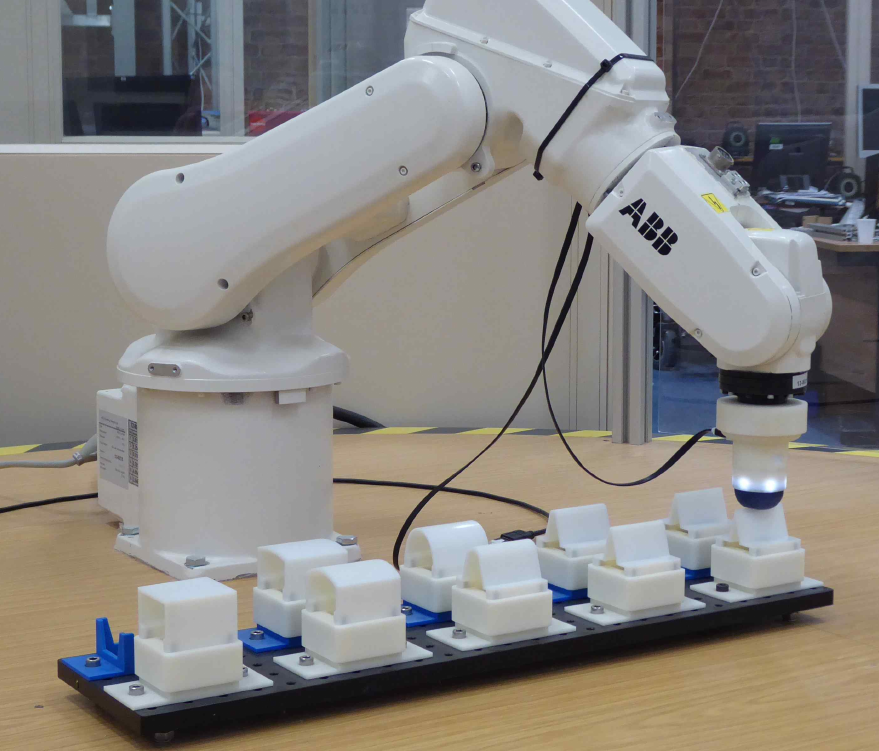}
    \includegraphics[width=0.36\linewidth,trim={255 27 0 0},clip]{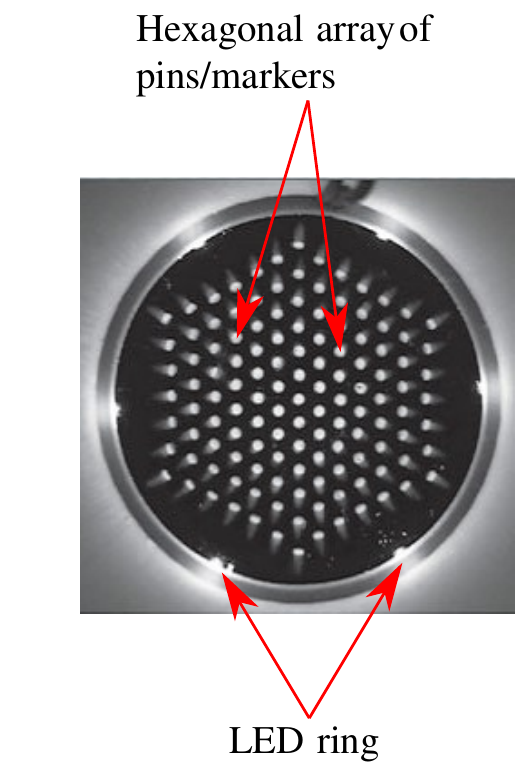}
    \includegraphics[width=\linewidth,trim={0 10 0 0},clip]{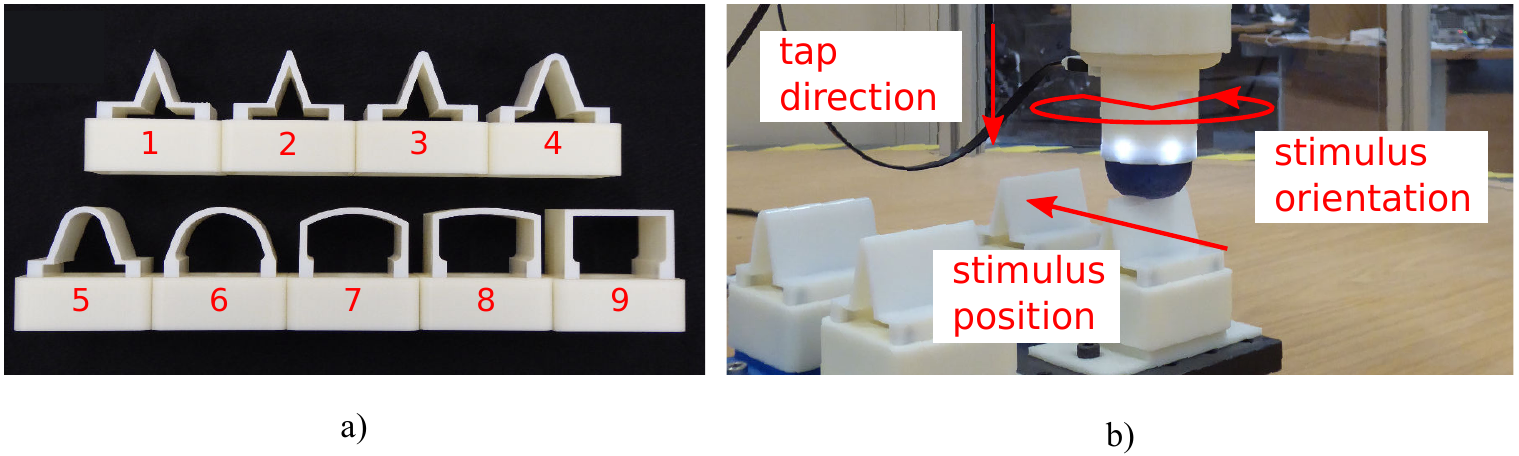}   
    \vspace{-2em}
    \caption{Robotic system used to capture the tactile data. The TacTip sensor is mounted as an end-effector on an IRB 120 robot arm and used to record tactile data from a set of 3D-printed stimuli mounted on a fixed plate. A hexagonal array of pins on the inside surface of the tip transduces and amplifies deformation of the skin surface. The nine edge-based stimuli span a range of curvatures, varying from a sharp edge to a flat surface.    }
    \vspace{-0em}
  \label{fig:robot_and_stimulus}
\end{figure}

%--------------------------------------------------------    

\subsubsection{Robot Arm} \label{subsubsec:robot_arm}
The TacTip was mounted as an end-effector on a six degree-of-freedom (DOF) IRB 120 robotic arm that can position the sensor with an absolute repeatability of approximately 0.1\,mm (Fig. \ref{fig:robot_and_stimulus}).

\subsubsection{Integrated Tactile Sensing and Robot Control} \label{subsubsec:software}
A modular software framework was used to control the robot arm and capture data from the tactile sensor. All high-level control and data processing was implemented using MATLAB. The robot arm was controlled using a Python client that communicates with a RAPID server running on the robot controller. The tactile sensor interface also performs several low-level image-processing functions for detecting and tracking the markers in the sensor image. 

\subsection{Dataset Collection and Preprocessing} \label{subsubsec:stimuli_and_colletction}
This study focuses on tactile perception of position, orientation and curvature of ridged stimuli (Fig~\ref{fig:robot_and_stimulus}, bottom left). We also aim to generalize across different combinations of these properties so that, for example, we can predict the orientation of a stimulus across a range of positions and curvatures. For these experiments, we used a set of nine 3D-printed ridged stimuli that span a range of curvatures. The curvature of each stimulus was represented using the Logarithm of the Radius of Curvature (LROC). Except for the 1st and 9th stimuli, the radius of curvature of all other stimuli increases by a factor of two across the range, and so the LROC scales results in a linear prediction scale. 

The data sets were generated by uniform random sampling 1,000 points over the ±15 mm range of positions and ±90-degree range of orientations for each stimulus.  For taps within this position range, all stimuli contacted the sensor in the ±10\,mm range, but only Stimuli 5-9 contacted the sensor between -10\,mm and -15\,mm, and between 10\,mm and 15\,mm.  This process resulted in datasets containing 9,000 samples. Two separate datasets were gathered: the first for training and hyperparameter optimization and the second for testing.  For each tactile sensing sample, the sensor was positioned approximately 1\,mm above the stimulus and then tapped down 5\,mm. Then five consecutive frames were recorded before returning the sensor to the start position. Each set of frames was labelled with the stimulus position, orientation and curvature.

%-------------------------------------------------------- 

Before being used as an input to a prediction model, each sensor frame was pre-processed in OpenCV, using an adaptive threshold and blob detection routine to produce a 254-element vector of $x$ and $y$ marker coordinates. A nearest-neighbour tracking routine was also used to ensure that the ordering of the 254-element vector remained consistent across datasets. The initial marker positions, recorded when the sensor was not in contact, were subtracted from the marker positions returned by the sensor to produce a vector of relative marker displacements.  The five samples collected during each tap were then averaged to produce a single 254-element observation vector. We found that collecting five samples and averaging them in this way helps to remove some of the noise associated with image capture and processing.

To investigate the impact of perceptual aliasing on performance, we generated two additional train/test pairs of datasets from the original pair.  The first additional pair was generated by filtering out the samples associated with flat Stimulus 9 and all samples lying outside the ±10 mm position range.  This pair of filtered datasets did not include any of the potential aliasing effects associated with indistinguishable free-space taps on either side of Stimuli 1-4, or indistinguishable taps across the flat Stimulus 9, and was used to exclude aliasing effects for position and orientation prediction.

The second additional pair of datasets was generated by filtering out the samples for Stimuli 1-3 and all taps lying outside the ±5 mm position range.  This pair of filtered datasets did not include any of the potential aliasing effects due to the TacTip not being sufficiently compliant to discriminate between the sharper edges of Stimuli 1-3, or it being unable to discriminate between curvatures in positions where there is minimal contact with the sensor.  This second additional pair of datasets was used to exclude aliasing effects for curvature prediction.  In this study we refer to the original pair of datasets as the \textit{aliasing} datasets, and the additional filtered pairs of datasets as the \textit{no-aliasing} datasets.

\begin{table*} [t!]
\centering
    \caption{Model performance (10 training runs): Mean and standard deviation RMS test errors}
    %\vspace{-1em}
    \label{tab:table_5}
    %\\ 
    \includegraphics[width=.8\linewidth]{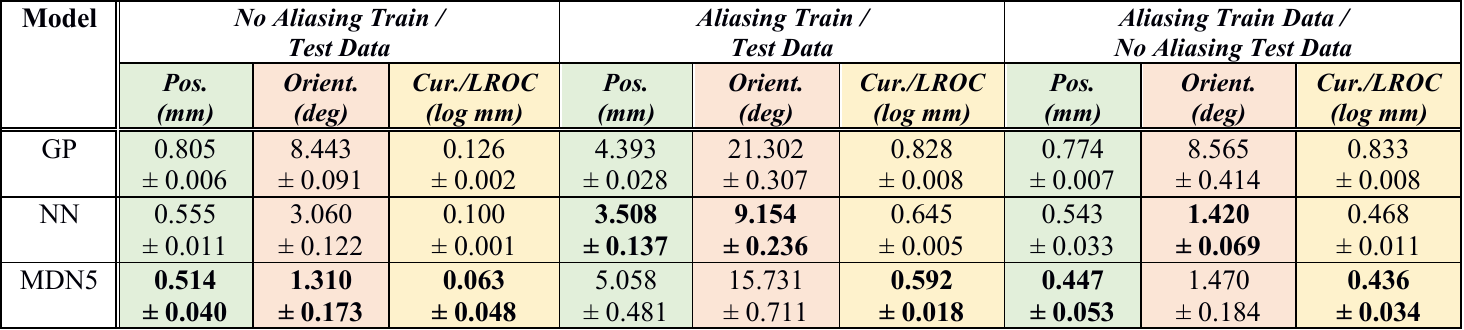}
    \vspace{-0em}
    % (b) Mean and standard deviation train and test times\\ 
    % \label{tab:table_6}
    % \includegraphics[width=1\linewidth]{tables/table_6_merged.pdf}
\end{table*}

\begin{figure*}[t!]
  \centering
    \includegraphics[width=0.86\linewidth]{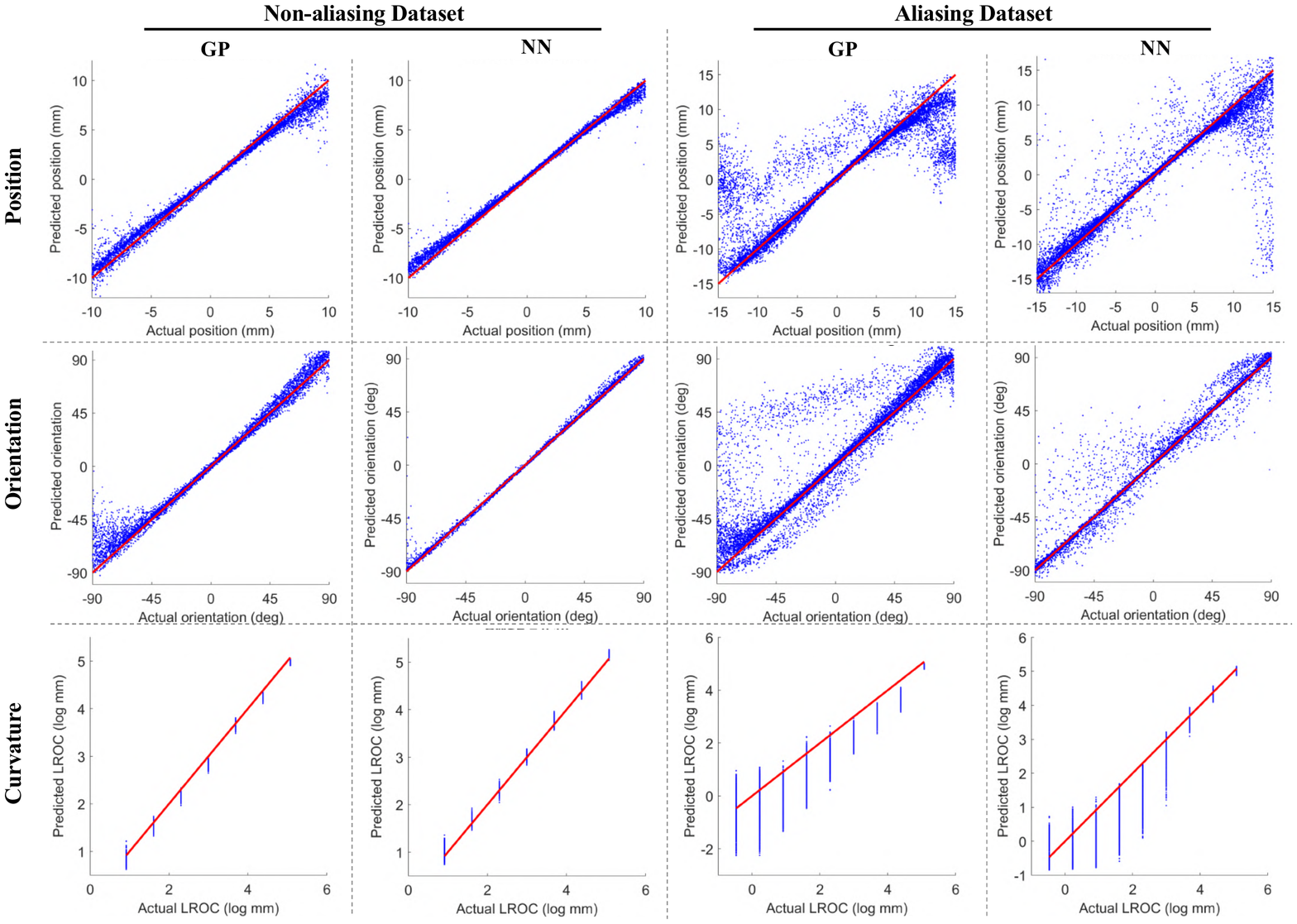}
    \caption{Results for the Gaussian process and deep neural network models trained on the no-aliasing and aliasing datasets. Predictions for position, orientation and curvature are plotted against the ground truth labels. Notice the significant scatter in the aliasing dataset with associated RMS errors in Table~2.}
  \label{fig:GP_NN}
% \end{figure*}
% \begin{figure*}[t!]
%  \centering
\vspace{1em}
    \includegraphics[width=.90\linewidth]{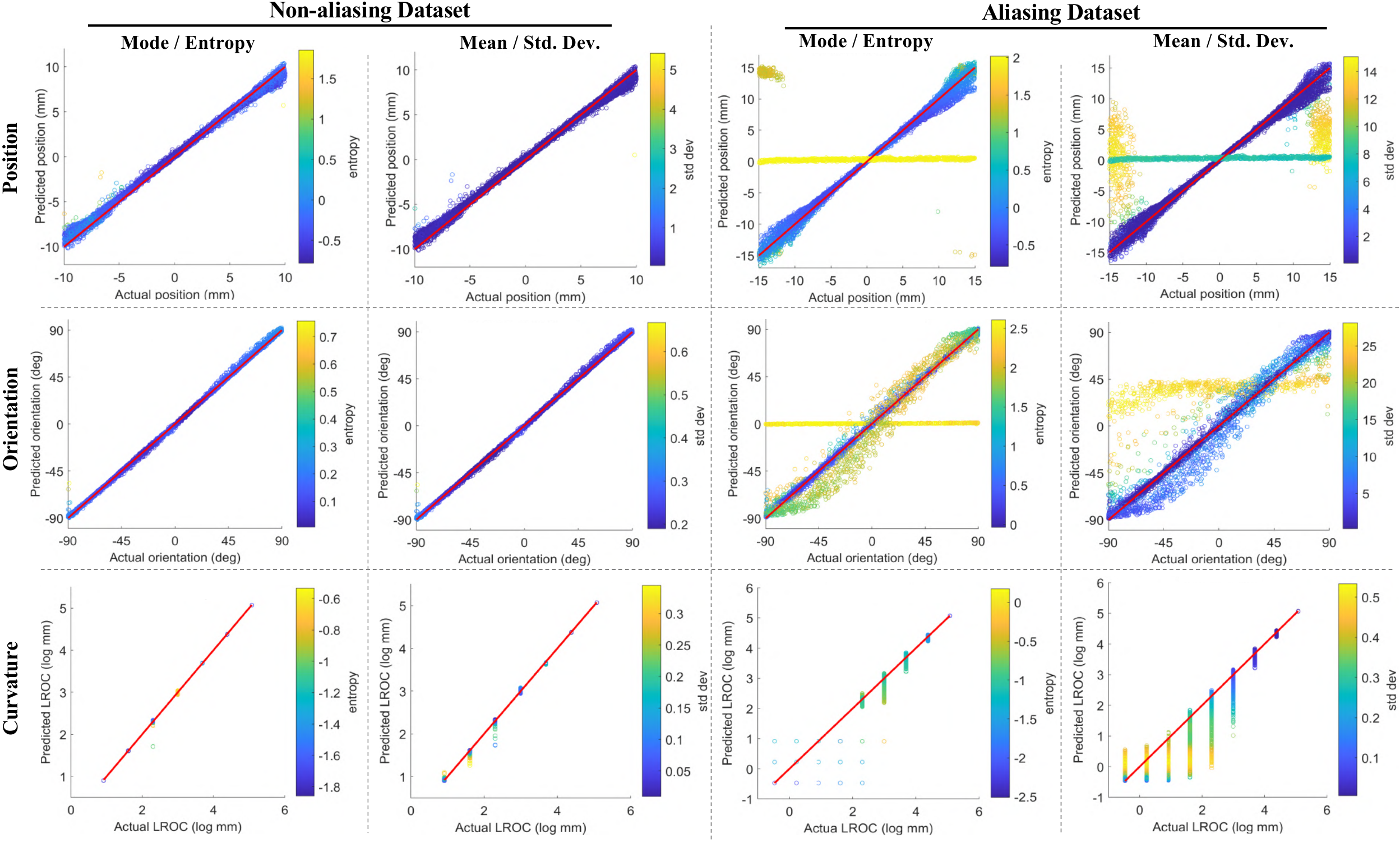}
    \caption{Results for the mixture density network model trained on the no-aliasing and aliasing datasets as in Figure~4. The prediction uncertainty is indicated by the colours of the points. Notice, again, how the high degree of scatter present in the aliasing dataset with associated RMS errors in Table~2.}
  \label{fig:uncertainty}
\end{figure*}

\section{Results} \label{sec:results}

\subsection{Predictive accuracy of discriminative regression models} \label{subsec:exp_1}

Two types of discriminative regression model were trained to predict the position, orientation and curvature of stimuli, based on the tactile sensor observations: Gaussian processes and deep neural networks. In each case, the model hyperparameters were optimized as described in the Methods (see Appendix, Table~\ref{tab:table_4}a,b). The model performance is summarized by the mean and standard deviation of the RMS test errors (Table~\ref{tab:table_5}).

The predictive accuracy of both models is significantly degraded when perceptual aliasing is present in the data (Fig.~\ref{fig:GP_NN}, {\em c.f.} left and right pairs of columns).  This is indicated by the larger RMS errors and wider dispersion of predictions around the red ground-truth line for the aliasing dataset when compared to the no-aliasing dataset, which is confirmed in the overall RMS test errors (Table~\ref{tab:table_5}). It is also clear that the NN models produce a lower RMS error and narrower distribution around the ground-truth line than the GP models, so they are more accurate for this set of tasks.  Note also that both models are only capable of producing single-point predictions and do not estimate prediction uncertainties (although a standard GP can estimate the predictive variance of an output variable, this only represents how confident it is about its prediction of the output; it does not model the distribution of noise around the output).

\subsection{Predictive accuracy of probabilistic discriminitive models versus discriminative regression models} \label{subsec:exp_2}

A similar analysis to that described in Section~\ref{subsec:exp_1} was applied to a 5-component mixture density network (MDN5), as an example of a probabilistic discriminative model. As above, the model hyperparameters were optimized as described in the Methods (see Appendix, Table~\ref{tab:table_4}c). For all MDN models, the predictions of position, orientation and curvature are all based on the location of the (major) mode of the predicted output distribution (see Appendix~\ref{subsec:dis_reg_model}). The model performance is again summarized by the mean and standard deviations of its RMS test errors over ten runs (Table \ref{tab:table_5}). 

The predictive accuracy of the MDN5 model was compared to the GP and NN models on the aliasing and no-aliasing tactile datasets (Table \ref{tab:table_5}). For each training run on the no-aliasing dataset, the RMS error was computed using the no-aliasing test set. For each training run on the aliasing dataset, the RMS error was computed using {\em both} the aliasing test set and the no-aliasing test set. The error on the no-aliasing test set indicates the residual error that would occur if it were possible to identify and remove the points associated with aliasing at test time.

For the no-aliasing data set, the MDN5 model is consistently more accurate than the other models (Table \ref{tab:table_5}, left 3 columns). The accuracy of all three models is generally better for no-aliasing data than it is for aliasing data.

For the aliasing data set, the NN model is the most accurate for position and orientation, and the MDN5 model is the most accurate for curvature (Table \ref{tab:table_5}, middle 3 columns). 

For the models trained using the aliasing training set and tested using the no-aliasing test set, the MDN5 models are consistently more accurate than the corresponding NN or GP models
(Table \ref{tab:table_5}, right 3 columns), except for orientation where the NN and MDN5 models are about the same.

\subsection{Single-point uncertainty values using probabilistic discriminative models} \label{subsec:exp_3}

One of the benefits of using an MDN model is that it is relatively easy to compute the uncertainty associated with predictions without resorting to computationally-expensive techniques such as Monte Carlo integration or optimization. More specifically, single-point predictions can be computed (or easily approximated) from either the mean, median or mode extracted from the Gaussian mixture model, and uncertainty can be computed using the standard deviation or entropy (see Appendix~\ref{subsec:dis_reg_model}).

Two prediction/uncertainty combinations were considered for the MDN5 model (mode/entropy and mean/standard deviation) on both the aliasing and no-aliasing data sets (Fig.~\ref{fig:uncertainty}; colour indicates the level of uncertainty). In general, the uncertainty is high where predictions are further from the red ground truth line or where the spread of predictions is larger.

For the no-aliasing dataset, there is relatively little variation in prediction accuracy, with just a few high-uncertainty outliers that lie some distance from the ground-truth line (Fig. \ref{fig:uncertainty}, left two columns). Note that the MDN model has captured the extremes of the position and orientation ranges better than the GP and NN models, with less scatter in the predictions. 

For the aliasing dataset, there is a striking variation in prediction accuracy originating from several different perceptual aliasing effects (Fig. \ref{fig:uncertainty}, right two columns):\\
\noindent (i) The central horizontal band for the position model (top row) corresponds to indistinguishable observations across the flat surface of Stimulus 9 where all locations 'feel' the same. Note how the predictions have regressed to the mean of 0\,mm.\\
\noindent (ii) The broad diagonal band of uncertain predictions around the ground truth line for orientation (middle row) appears due to the difficulty of perceiving the rotation of the flatter stimuli. \\
\noindent (iii) The near-central horizontal band for the orientation model (middle row) corresponds to indistinguishable observations when the sensor is close to tapping in free space and makes little or no contact with the sharper Stimuli 1-4.\\ % It also has a contribution from the flat Stimulus 9.\\
\noindent (iv) The clusters at the extremes of the position range (top row) correspond to those same indistinguishable observations when the sensor makes little or no contact with Stimuli 1-4. \\ 
\noindent (v) The curvature model (bottom row) has worsening predictive accuracy and increasing uncertainty for the sharper Stimuli 1-4 as above due to little or not contact. This ambiguity is amplified by the sensor skin not being sufficiently flexible to conform to the sharpest ridged stimuli.

Additionally, we note that the entropy measure appears to distinguish between high-uncertainty perceptual aliasing contacts and low-uncertainty conventional contacts to a far greater extent than the standard deviation. In particular, the visual separation between the two classes of tactile contact is much clearer when using the mode and entropy measure than when using the mean and standard deviation.

\subsection{Application to sequential decision making} \label{subsec:exp_4}

\begin{figure}[t]
  \centering
    \includegraphics[width=.49\linewidth,trim={0 0 330 0},clip]{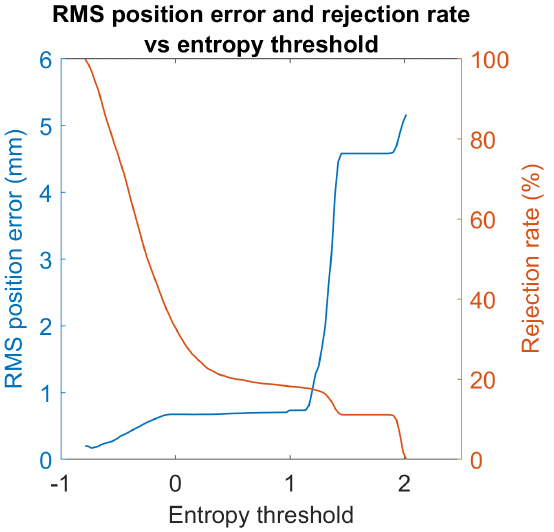}
    \includegraphics[width=.49\linewidth,trim={330 0 0 0},clip]{pics/mdn_paper_fig_7.pdf}
    \vspace{-2em}
    \caption{Sequential decision making where samples are rejected if they are above an uncertainty threshold. The MDN5 test error and rejection rate are plotted against an entropy threshold for the position and orientation models. The error improves as more samples are rejected with lower uncertainties.}
    \vspace{-0.7em}
  \label{fig:RMS}
\end{figure}

To illustrate one benefit of uncertainty modelling for tactile perception, we consider the use of the MDN probabilistic discriminative model for a sequential decision making (SDM) scenario, where a sufficiently accurate decision must be made with as few sequential observations as possible. 

A typical (normative) framework for optimal decision-making uses a cost function that depends on both the decision error and decision time (number of observations). An optimal solution to this type of problem  takes the form of a {\em stopping rule} that accepts a prediction if the uncertainty is lower than a specified threshold but otherwise gathers more samples until the threshold is reached. The threshold is usually chosen or optimized to balance the cost of making errors against the cost of further sampling: a lower threshold tends to decrease the final decision error but leads to an increase in the number of rejected samples, and {\em vice versa}~\cite{wald_sequential_1947}.

Since an MDN model can be used to make predictions and estimate the associated uncertainty, it can be used to investigate how the overall error and rejection rate varies for different uncertainty thresholds. This is illustrated for position and orientation SDM prediction tasks in Fig.~\ref{fig:RMS}, assuming the curvature does not change if a sample is rejected. 

For both perceived quantities, the overall trend is that increasing the entropy-based uncertainty threshold results in an increase in RMS decision error, accompanied by a decrease in sample rejection rate, as expected (Fig. \ref{fig:RMS}).

An interesting feature is that both the position and orientation models have a range of thresholds where there is both relatively low RMS error and a low rejection rate ({\em i.e.} a `sweet spot'). These threshold ranges lie between approximately 0.0 to 1.0 (left panel) and 0.5 to 1.5 (right panel), where the RMS errors are approximately 0.7\,mm for position and 1\,deg for orientation, upon rejecting about 1 in 10 samples. These accuracies are comparable with those from the non-aliasing test set (Table~\ref{tab:table_5}), by using the model to reject ambiguous samples rather than tuning the parameter ranges. 

These results could be improved further with a Bayesian approach that fuses predictions and their uncertainties over sequential observations, and with more intelligent policies for controlling the sensor to actively sample data.

%----------------------------------------------------------------------------------
\section{DISCUSSION} \label{sec:discussion}
%--------------------------------------------------------------------------------
This work shows that perceptual aliasing can be a serious issue for discriminative prediction methods such as Gaussian process regression and deep neural networks that are in common use in artificial tactile sensing. One solution is to remove the sources of aliasing, which results in improved predictive performance; however, this requires human intervention to fine-tune the data collection. We have shown instead that a probabilistic discriminative model (a mixture density network) can address this tactile perceptual aliasing problem, by modelling the distribution of the data so that the aliasing is captured. The probabilistic model gives both a prediction and its uncertainty, which can be used to identify aliased data. 

We view the tactile aliasing problem as primarily one of data uncertainty, where labelled data is available to construct a predictive model, but that data has a distinctly non-Gaussian or multi-modal distribution.  This contrasts with the complementary problem of model uncertainty, which arises when some of the data is missing or unavailable over part of the domain. 

The ability of the mixture density network to model perceptual aliasing enables it to properly capture the distribution of tactile data. This is evident from the test results that reveal structures in the distribution, which are coloured according to the level of uncertainty (Fig.~\ref{fig:uncertainty}, right columns). In addition to being located away from the ground truth line, the uncertain predictions form patterns that are consistent with the various sources of perceptual ambiguity; for example, clusters at extreme positions for light contacts, and horizontal bands of unpredictable positions for the flat stimulus. These structures are absent from the results for the two discriminative models (Fig,~\ref{fig:GP_NN}, right two columns) because they cannot model the multi-modal, distributed nature of the perceptual aliasing.  

While it is recognized that standard deep neural networks are not uncertainty-aware, one might consider that Gaussian process regression provides a useful measure of uncertainty. However, standard Gaussian processes only estimate the predictive variance, which is a measure of uncertainty in the prediction of the conditional mean, not an estimate of the conditional uncertainty in the output data distribution (they use a single global hyperparameter for the output data noise, which is optimized during model fitting). To describe perceptual aliasing, the output data uncertainty would need to be conditioned on the input variables. Heteroscedastic Gaussian processes that allow for input-dependant noise have been considered~\cite{le_heteroscedastic_2005,kersting_most_2007}; however, we are not aware of how those methods could be applied to heavy-tailed or multi-modal distributions which were necessary here to model tactile perceptual aliasing.

Uncertainty-aware deep learning is recognized as an important topic for robot vision. Recent work by Loquercio {\em et al} introduced a general framework for uncertainty estimation in deep learning~\cite{loquercio_general_2020}. Their model is based on a Bayesian belief network~\cite{frey_variational_1999} with Monte Carlo sampling of predictions implemented using dropout~\cite{gal_dropout_2016}. While their approach captures both model and data uncertainty to some extent, it relies on propagating single-point estimates of uncertainty (assumed Gaussian at the inputs) through the model to compute single-point estimates of the mean output predictions and their uncertainties, rather than providing a full conditional distribution of the model outputs. Using that approach, it would not be possible to distinguish multi-modal distributions from uni-modal distributions having the same mean and variance.

How would the uncertainty estimate be used in practice? We considered rejecting samples in a sequential decision making task if the uncertainty was too high, then re-sampling at a random location (Fig.~\ref{fig:RMS}). A better sampling policy would choose an action that addresses the tactile aliasing, based on knowledge contained in the model of how the uncertainty depends on sampling location relative to the object. Other information such as the sensor pose could be used in this policy, although pose would not be useful on the first contact if the object or its pose are unknown in advance. Future work in this area could relate tactile perceptual aliasing to established approaches in haptics such as active perception~\cite{bajcsy_active_1988}.

To conclude, the use of deep neural networks and other discriminative models bring many advantages for tactile robotic applications. However, those models are not able to capture distributed input data under tactile perceptual aliasing. In our view, perceptual aliasing will become an unavoidable issue for robot touch as the field progresses to gather training data from robots acting in uncertain and unstructured environments. Deep reinforcement learning for robot manipulation is one area where perceptual aliasing could present significant difficulties, as tactile data is more commonly adopted as a principal modality for controlling physical interaction.  

\vfill\pagebreak

%\cite{ITER_supplement} 

% \section{CONCLUSION} \label{sec:conclusion}
% %--------------------------------------------------------------------------------

%----------------------------------------------------------------------------------
% \section*{ACKNOWLEDGMENT}
% %----------------------------------------------------------------------------------
% The authors thank members of the BRL Tactile Robotics group for help and advice. JL and NL were supported by an award from the Leverhulme Trust on `A biomimetic forebrain for robot touch' (RL-2016-39)
%\vfill\pagebreak

%% BIBLIOGRAPHY
%----------------------------------------------------------------------------------
\bibliographystyle{IEEEtran}
\bibliography{IEEEabrv,references,manual}
%------------------------------------------------------------------------------------------------------------------

\vfill\pagebreak

\appendix

%----------------------------------------------------------------------------------
\subsection{Hyperparameter optimization} \label{sec:back}

\begin{table} [h!]
\centering 
\vspace{-1em}
    \caption{Optimized hyperparameter values}
    %\label{tab:table_2}
    \vspace{-1em}
    (a) Gaussian process models\\
    \includegraphics[width=\linewidth]{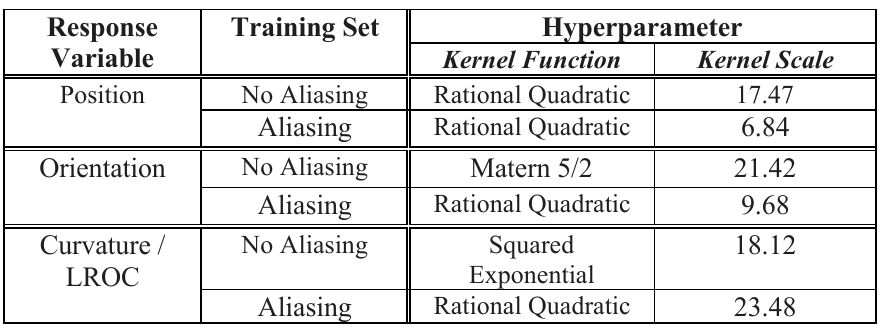}
    (b) Neural network models\\
    %\label{tab:table_3}
    \includegraphics[width=\linewidth]{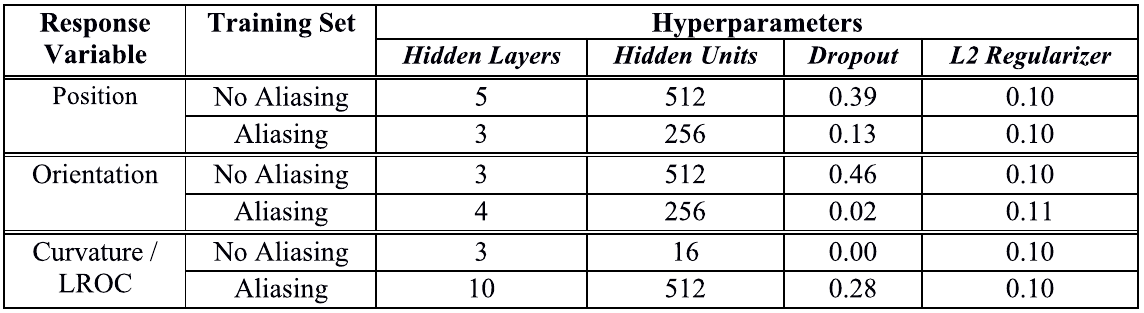}
    (c) Mixture density network model with five components
    \label{tab:table_4}
    \includegraphics[width=\linewidth]{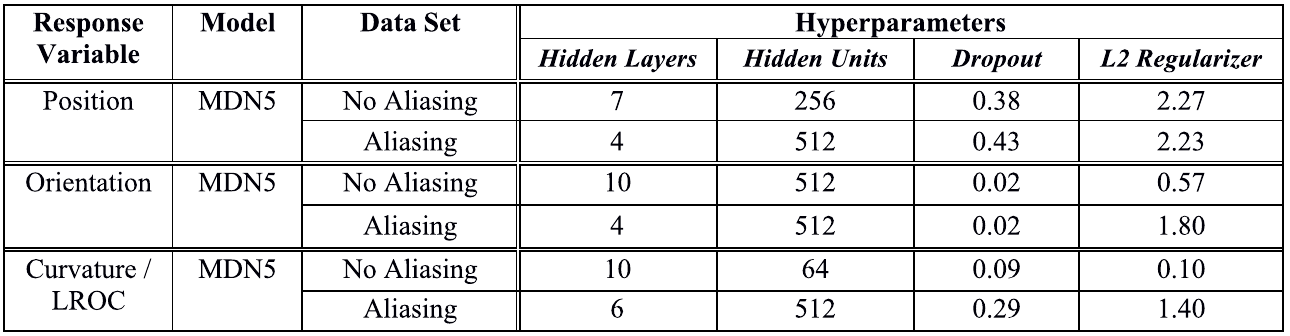}
\vspace{-1em}
\end{table}

%----------------------------------------------------------------------------------
%\subsection{Appendix B} \label{sec:back}
%----------------------------------------------------------------------------------

%----------------------------------------------------------------------------------
\subsection{Probabilistic model of tactile stimulation and perception}\label{subsec:tactile_prob_model}
%----------------------------------------------------------------------------------
In this paper, we model the output of the tactile sensor as a continuous random vector $\boldsymbol{x}=(x_1,x_2,...,x_M) $ that represents the output of individual tactile elements, or `taxels'. We also model a tactile stimulus as a continuous random vector $\boldsymbol{y}=(y_1,y_2,...,y_N) $ that represents attributes such as position, orientation and curvature.
The probabilistic relationship between stimulus and sensor output is governed by a joint PDF $p(\boldsymbol{x},\boldsymbol{y})$. Thus, the causal forward process of transducing a stimulus into a tactile sensor output is modelled by the conditional PDF $p(\boldsymbol{x}|\boldsymbol{y})$.  Similarly, the inverse inference process of tactile perception, where the applied stimulus that gave rise to an observed sensor output is inferred, is modelled by the conditional PDF $p(\boldsymbol{y}|\boldsymbol{x})$. 

Following this line of reasoning, in the absence of noise we assume that the tactile sensor output $\boldsymbol{x} $ can be expressed as a deterministic, nonlinear function $\boldsymbol{x} = g(\boldsymbol{y})$ of the applied stimulus.  However, due to limited accuracy, both stimulus and output may be corrupted by additive noise sources:
\begin{align}\label{eq:forward_with_noise}
    \boldsymbol{x}=g(\boldsymbol{y}+\boldsymbol{\epsilon_y})+\boldsymbol{\epsilon}_x.
\end{align}

For simplicity, we assume the noises $\boldsymbol{\epsilon}_x$ and $\boldsymbol{\epsilon}_y$ are sampled from normal distributions $\mathcal{N}(\boldsymbol{0},\boldsymbol{\Sigma}_x)$ and $\mathcal{N}(\boldsymbol{0},\boldsymbol{\Sigma}_y)$ with zero means and diagonal covariance matrices $\boldsymbol{\Sigma}_x$ and $\boldsymbol{\Sigma}_y$.

In tactile perception, we are usually more interested in the inverse model, which helps us infer the stimulus $\boldsymbol{y}$ that was applied to generate an observed sensor output $\boldsymbol{x}$:
\begin{align}\label{eq:backward_with_noise}
    \boldsymbol{y}=g^{-1}(\boldsymbol{x}+\boldsymbol{\epsilon}'_{x})+\boldsymbol{\epsilon}'_{y},
\end{align}
where $\boldsymbol{\epsilon}'_{x} = -\boldsymbol{\epsilon}_{x}$ , and $\boldsymbol{\epsilon}'_{y} = -\boldsymbol{\epsilon}_{y}$. 

Two complications can arise when using this inverse model.  First, if two or more stimuli produce the same sensor output, then the sensor transformation $g(\cdot)$ might not be invertible; {\em i.e.} there is perceptual aliasing, which can give rise to multimodality in the conditional distribution of $\boldsymbol{y}$. Second, even if the sensor transformation is invertible and the sensor output noise $\boldsymbol{\epsilon}'_{x}$ is i.i.d. Gaussian, after the noise is referred through the nonlinear function $g^{-1}(\cdot)$ its distribution will change, and the components may become dependent, asymmetric, heteroscedastic or multimodal. This problem becomes more significant where higher levels of sensor output noise $\boldsymbol{\epsilon}'_{x}$ are present. Therefore, an important criterion for selecting a suitable tactile perception model is whether it has sufficient representational power to capture the features of the distribution implied by Equation~(\ref{eq:backward_with_noise}).

%----------------------------------------------------------------------------------
\subsection{Discriminative regression models}\label{subsec:dis_regre_model}
%----------------------------------------------------------------------------------
In terms of the tactile perception (inference) problem, discriminative regression models such as Gaussian processes and neural networks assume an underlying model of the form $\boldsymbol{y}=f(\boldsymbol{x})+\boldsymbol{\epsilon}_y$, with noise $\boldsymbol{\epsilon}_y\sim \mathcal{N}(\boldsymbol{0}, \boldsymbol{\Sigma}_Y)$. Comparing this with the model in Equation~(2), we see that it implicitly assumes that: (a) the sensor transformation $g(\cdot)$ is invertible, and (b) the sensor output noise $\boldsymbol{\epsilon}_x$ is zero or negligible.  If either assumption does not hold, the predictive accuracy of the model will be significantly degraded.

{\em Gaussian Processes}\label{subsubsec:GP} are a prior distribution over real-valued functions, $f\sim \mathcal{GP}(m,k)$. The prior enforces constraints on function attributes such as smoothness and amplitude. The functions $m(\boldsymbol{x})$ and $k(\boldsymbol{x},\boldsymbol{x}')$ are the prior mean and covariance (kernel) functions. The squared exponential kernel is one of the most popular types of kernel: $k(\boldsymbol{x},\boldsymbol{x}')=\sigma_{s}^{2}exp(-(\boldsymbol{x}-\boldsymbol{x}')^{2})/2l^{2})$, where the signal level $\sigma_s$ and length scale $l$ are hyperparameters. 

% The GP posterior
Once some data $\mathcal{D}$ has been observed, the prior can be replaced by a posterior distribution over functions $f\mid \mathcal{D} \sim \mathcal{GP}(m_\mathcal{D},k_\mathcal{D})$, which can be used to predict values for new test points.  Here, $m_{\mathcal{D}}(\boldsymbol{x})$ and $k_{\mathcal{D}}(\boldsymbol{x},\boldsymbol{x'})$ are the posterior mean and covariance functions corresponding to those for the prior. 

One of the problems with GPs is that they do not tend to scale well to larger datasets. This has led to the development of several sparse approximation methods that can cope with larger datasets at the expense of lower predictive accuracy.

{\em Neural Networks}\label{subsubsec:NN} are parameterised models of the form $\boldsymbol{\hat{y}}=f(\boldsymbol{x};\boldsymbol{\varphi})$. The set of weight matrices and bias vectors constitute the model parameters~$\boldsymbol{\varphi}$. The parameters are updated in an iterative manner by following the negative gradient of an error function $E(\cdot)$ that measures how close the predicted outputs $\left \{ \boldsymbol{\hat{y}}_{i} \right \}$ are to a set of target outputs $\left \{ \boldsymbol{y}_{i} \right \}$ for the inputs  in a training data set $\mathcal{D} = \left \{ \boldsymbol{x}_{i},\boldsymbol{y}_{i} \right \}$. For regression problems, the error function  is usually taken to be the mean-squared error between the model outputs and targets or, equivalently, the negative conditional log-likelihood of the training data $\mathcal{D}$ under an assumed Gaussian distribution.

% Hyperparameters
Within this general framework there is still a high degree of flexibility in how the model can be configured: number of hidden layers, size of each layer (number of layer outputs), activation function, optimizer type and configuration, normalization methods, regularization methods, and so on.

\subsection{Discriminative distribution models}\label{subsec:dis_reg_model}
Unlike discriminative regression models, probabilistic discriminative models do not make strong assumptions about the underlying probabilistic model $p(\boldsymbol{y}|\boldsymbol{x})$.  In fact, assuming they have sufficient representational capability, they are capable of representing any conditional PDF to an arbitrary accuracy.  So, in principle, they can represent models of the form specified in Equation~(\ref{eq:backward_with_noise}), regardless of whether the sensor transformation is invertible or the sensor output noise  is negligible.  However, since fitting a full distribution model generally requires more data than to fit simpler statistical functions like the mean, this increased flexibility comes at a price.

{\em Mixture Density Networks}\label{subsubsec:MDN} combine the distribution modelling flexibility of a mixture model, with the parameter prediction (conditioning) power of a neural network.  Although various types of mixture model can be used, in this paper we restrict our attention to finite mixtures of $K$ multivariate Gaussian components:
\begin{align}\label{eq:GMM}
    p(\boldsymbol{y})=\sum_{i=1}^{K} \alpha_i\mathcal{N} (\boldsymbol{y};\boldsymbol{\mu}_i,\boldsymbol{ \Sigma}_i).
\end{align}
The parameters  $(\alpha_1, \alpha_2, ..., \alpha_K)$ are the mixture weights, which are positive and sum to one; $(\boldsymbol{\mu}_1, \boldsymbol{\mu}_2, ..., \boldsymbol{\mu}_K)$ are the component means, and $(\boldsymbol{\Sigma}_1, \boldsymbol{\Sigma}_2, ..., \boldsymbol{\Sigma}_K)$ are the component covariances.   The mixture weights, component means and covariances are combined/flattened into a parameter vector $\boldsymbol{\theta}$, and predicted by the outputs of a multilayer neural network with parameters $\boldsymbol{\varphi}$: $\boldsymbol{\theta}=f(\boldsymbol{x};\boldsymbol{\varphi})$. This provides the conditioning mechanism for expressing the probability distribution $p(\boldsymbol{y}|\boldsymbol{x})$ of outputs $\boldsymbol{y}$ as a function of the inputs $\boldsymbol{x}$. 

The model is trained by minimizing the negative conditional log-likelihood of the training data $\mathcal{D}={(\boldsymbol{x}_i,\boldsymbol{y}_i)}$ under an assumed Gaussian mixture distribution.  Since this corresponds to a maximum likelihood (ML) approach, it is common to add a regularization penalty to the log-likelihood error to help prevent over-fitting. The minimization is carried out using the same type of stochastic gradient descent-based optimizer that is used for conventional neural networks.

Once the MDN has been trained, the mixture model parameters can be used to make predictions and estimate uncertainty.  In this study, we make predictions using the conditional mean or mode of the predicted distribution.  We calculate the \emph{conditional mean} as:
\begin{align}\label{eq:GMM_mean}
    \boldsymbol{\mu}=\sum_{i=1}^{K} \alpha_i \boldsymbol{\mu}_{i} .
\end{align}
Unfortunately, it is not possible to express the conditional (major) mode using a closed-form analytical expression.  So, rather than use slower numerical methods to solve for this value, we use a faster closed-form approximation based on the mean of the highest density component.  Hence, we approximate the \emph{conditional mode} as:
\begin{align}\label{eq:GMM_cond_entropy}
    \boldsymbol{M}=\boldsymbol{\mu}_k,\ \ \  k=\underset{i}{\rm argmax}\left \{ \frac{\alpha_i}{ \sqrt{{\rm det}(\boldsymbol{\Sigma}_i)} } \right \}.
\end{align}
This provides a reasonably good approximation where the components do not overlap too much. However, for other tactile perception tasks, it might make more sense to use the mean of the most probable mixture component to make predictions:
\begin{align}\label{eq:GMM_cond_entropy_approx}
    \boldsymbol{M}=\boldsymbol{\mu}_k,\ \ \  k=\underset{i}{\rm argmax}\left \{ \alpha_i \right \}.
\end{align}
We calculate the uncertainty associated with an MDN prediction using one of two methods: we either use the 'standard deviation', which we calculate as the square root of the trace of the conditional variance $S=\sqrt{{\rm tr}(\boldsymbol{ \Sigma})}$; or the entropy.  The \emph{conditional variance} is given by:
\begin{align}\label{eq:GMM_cond_var}
    \boldsymbol{ \Sigma}=\sum_{i=1}^{K} \alpha_i \boldsymbol{ \Sigma}_i + \sum_{i=1}^{K} \alpha_i (\boldsymbol{\mu}_i - \boldsymbol{\mu}) (\boldsymbol{\mu}_i - \boldsymbol{\mu}) ^{\rm T}.
\end{align}
The conditional variance $\boldsymbol{\Sigma}$ arises from two sources of variability: within-component heterogeneity associated with the first term in Equation~(7), and between-component heterogeneity associated with the second.

We approximate the \emph{conditional entropy} using a component-wise, first-order Taylor-series expansion of the
logarithm of the Gaussian mixture~\cite{huber2008entropy}:
\begin{align}\label{eq:GMM_entropy_approx}
    H=-\sum_{i=1}^{K}\alpha_{i}\log\sum_{j=1}^{K}\alpha_j\mathcal{N}(\boldsymbol{\mu}_{i};\boldsymbol{\mu}_{j},\boldsymbol{ \Sigma}_{j}).
\end{align}

\end{document}